\documentclass[11pt, letterpaper]{gvcmix}

\usepackage[comma,numbers,sort,compress]{natbib}
\usepackage{url}
\usepackage{booktabs}
\usepackage{nicefrac}
\usepackage{microtype}
\usepackage{xcolor}

\usepackage{graphicx}
\usepackage{subcaption}
\usepackage{algorithm}
\usepackage[noEnd,commentColor=black]{algpseudocodex}
\usepackage[most]{tcolorbox}
\usepackage{float}
\usepackage{wrapfig}
\usepackage{lipsum}
\usepackage{colortbl}
\usepackage{soul}
\usepackage{enumitem}
\usepackage{multirow}
\usepackage{makecell}
\usepackage{tablefootnote}
\usepackage[para,online,flushleft]{threeparttable}
\usepackage[nameinlink,noabbrev]{cleveref}
\usepackage{siunitx}
\usepackage{tikz}
\usetikzlibrary{arrows.meta}

\definecolor{myblue}{rgb}{1,0.5,0}
\definecolor{urlteal}{HTML}{0F766E}
\definecolor{linkaccent}{HTML}{5A3CC4}
\definecolor{urlaccent}{HTML}{2A5FD8}
\hypersetup{
    colorlinks = true,
    linkcolor = {linkaccent},
    citecolor = {accent},
    urlcolor = {urlaccent},
    pdftitle = {Generative Value Classifier},
    pdfauthor = {Alexander Swerdlow, Perry Dong, and Chelsea Finn},
}
\renewcommand{\tt}[1]{\texttt{#1}}

\setlist{label=\textbullet}
\newcommand{\oursname}{GVC}

\newcommand{\oursplain}{{\texttt{\oursname}}}
\newcommand{\nonsubmissiononly}[1]{#1}

\newcommand{\codeurl}{https://github.com/alexanderswerdlow/gvc}
\newcommand{\projecturl}{https://alexanderswerdlow.github.io/gvc}



\title{\centering Generative Value Classifier}

\author{%
    \textbf{Alexander Swerdlow} \hspace{0.75em}
    \textbf{Perry Dong} \hspace{0.75em}
    \textbf{Chelsea Finn} \\\vskip 0pt
    {\normalfont \normalsize Stanford University}\vskip -1pt
}

\correspondingauthor={\footnotesize Correspondence to \texttt{aswerdlo@stanford.edu}}

\teaser{
    \centering
    \includegraphics[width=0.86\linewidth]{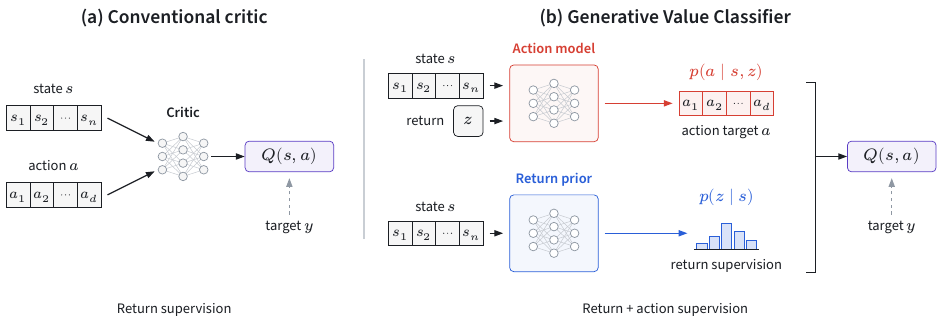}
    \captionsetup{hypcap=false}
    \captionof{figure}{\textbf{Action data provides additional supervision.} A conventional critic fits $Q(s,a)$ to a scalar target $y$. GVC instead learns a return-conditioned action model $p(a\mid s,z)$ and a return prior $p(z\mid s)$, i.e., a distributional state-value function, and normalizes their product over returns to obtain $Q(s,a)$.}
    \label{fig:teaser}
}

\begin{abstract}
Q-learning is premised on learning a function to predict the value of a state-action pair, with the goal of using such a function to guide policy improvement. Most approaches learn this function from scratch using scalar or distributional return targets. This discriminative approach leaves another source of supervision for learning representations unused: the actions themselves. Rather than directly predict the return of a state-action pair, we ask the reverse question: how likely is the action under each possible return? This perspective provides a critic whose generative component is a return-conditioned policy, which can share the actor's architecture. Training this policy supervises the critic with every action rather than with returns alone, and a squared-error loss anchors the critic's Q-value to the bootstrapped target. We evaluate our approach on both offline and online RL tasks and show gains over leading discriminative methods, including distributional RL methods.
\par\medskip
\noindent Code: \expandafter\url\expandafter{\codeurl}\\
Project page: \expandafter\url\expandafter{\projecturl}
\end{abstract}

\begin{document}

\maketitle

\section{Introduction} \label{sec:introduction}

Value functions play a key role in sample-efficient learning, enabling policy improvement to use past rollouts to better score newer rollouts. However, learning good value functions particularly in sparse-reward settings has remained a significant challenge~\citep{arjonamedina2019rudder,kumar2020discor}. Standard Monte-Carlo (MC) approaches result in high variance and lack the ability to stitch information across trajectories whereas Temporal-Difference (TD) bootstrapping addresses this problem but at the cost of compounding errors that propagate across timesteps. Sparse-reward settings exacerbate these issues with longer-horizon trajectories and necessitate additional rollouts to provide accurate value estimation.

Prior work has sought to improve value learning under sparse supervision through richer prediction tasks. Distributional RL~\citep{bellemare2017distributional} learns the return distribution rather than a single scalar. Other approaches use contrastive learning to predict future-state occupancy by contrasting known future states with noisy negatives~\citep{eysenbach2021clearning,eysenbach2022contrastive,mazoure2023contrastive}. These methods reformulate value learning through additional representational capacity or inductive biases; we instead ask whether modeling actions can supply additional supervision.

Our insight is that traditional value learning fails to take advantage of a key source of supervision: the actions themselves. Value functions are \textit{conditioned} on state-action pairs, yet are only supervised by a scalar---or at best distributional---return. To use the actions as supervision, we learn $p(a\mid s,z)$ rather than directly predicting $p(z\mid s,a)$, where $z$ denotes the return. With the state held fixed, Bayes' rule combines this likelihood with a return prior $p(z\mid s)$, the return distribution given the state alone. Normalizing their product over return bins gives a predicted return distribution whose mean is $Q(s,a)$. We also minimize the squared error between $Q(s,a)$ and the bootstrapped target return, which we call \textit{value anchoring}. This uses a discriminative loss to fit the value while retaining the generative supervision and structure.

In summary, our main contributions are:

\begin{itemize}[leftmargin=1.5em,labelsep=0.5em]
    \item We introduce a new formulation for Q-learning that leverages a return-conditioned generative likelihood to implicitly model the return distribution.
    \item We develop a hybrid generative--discriminative objective that combines return-conditioned flow matching with
    value anchoring. This aligns the return predictions with the standard Bellman backup targets while retaining the generative supervision.
    \item We evaluate \oursplain{} on challenging continuous-control tasks, including online RL benchmarks on RoboMimic and offline and offline-to-online RL benchmarks on OGBench, demonstrating strong performance relative to prior methods. Our analysis demonstrates that removing this generative supervision correlates with lower performance on the same architecture.
\end{itemize}

\section{Related Work}

\paragraph{Discriminative Tasks Using Generative Models.}
Using generative models for discriminative tasks has a long history, with~\cite{ngDiscriminativeVsGenerative2001} first comparing generative and discriminative classifiers and subsequent work learning a single model with both generative and discriminative objectives~\citep{lasserrePrincipledHybridsGenerative2006}. Driven by the strong performance of conditional generative models, recent work has built upon this, demonstrating that pretrained image and text models act as strong zero-shot classifiers, such as with text-to-image models~\citep{liYourDiffusionModel2023a,clarkTexttoImageDiffusionModels2023b} or with large language models~\citep{puri2019zeroshot,kumar2024genz}. Finally, another set of works, such as JEM~\citep{grathwohlYourClassifierSecretly2019} and HDGE~\citep{liuHybridDiscriminativeGenerative2020}, reinterprets standard discriminative classifiers as energy-based models (EBMs) and adds maximum likelihood or contrastive losses. Our work brings this view to reinforcement learning, where the ``class'' is a discretized return and the generative model is an action generation policy, thus constructing an implicit Q-function.

\paragraph{Generative Policies as Value Functions.}
\looseness=-1
At the same time, both reward modeling and value learning have made significant use of generative policies. In the language domain, reward models are commonly finetuned generative language models used to provide scalar rewards~\citep{ouyangTrainingLanguageModels2022}, or to provide text-based responses to grade actions based on a set of criteria~\citep{yuanSelfRewardingLanguageModels2024,zhangGenerativeVerifiersReward2024}. Similar methods have been applied to robotics, leveraging pretrained \textit{language} policies~\citep{ma2025gvl,chenTOPRewardTokenProbabilities2026,lee2026roboreward} for reward or value models, but these methods are unable to judge policy \textit{outputs}, i.e., actions, unlike their language domain counterparts. Finally, a recent line of work on RL for Vision-Language-Action (VLA) models has adopted the pretrained \textit{state} representation from policies as an initialization for value functions~\citep{zhai2025vlac,liu2025rlvla}. Similar to the language reward models above---though unlike these robotics methods---\oursplain{} scores the policy's actions directly. A robot policy (e.g., one that generates joint positions) has no direct way to output a value that is compatible with its pretraining, unlike a language model, which can simply state one in text. This motivates \oursplain{}: we instead condition a generative policy on the return and score each action by its likelihood under different returns.

\paragraph{Value Function Learning.} A separate line of work maintains standard value-based RL but increases the expressivity of the value function itself. Distributional RL models the full return distribution rather than its mean \citep{bellemare2017distributional}, with quantile regression \citep{dabney2017distributional} and implicit quantile networks \citep{dabney18implicit} increasing distributional flexibility. Recent work replaces these heads with stronger generative architectures for added expressivity \citep{dong2026valueflows, agrawalla2026flowmatching}, with Value Flows~\citep{dong2026valueflows} modeling the full return distribution with flow matching. Another set of works instead uses a transformer backbone with self-attention over state-action tokens in place of an MLP: Q-Transformer~\citep{chebotar2023qtransformer} exploits this to causally predict Q-values across action dimensions and timesteps, while PAC~\citep{springenberg2024offline} scales actor-critic training to large models; however, naively scaling such transformer critics induces attention entropy collapse, which TQL~\citep{dong2026tql} addresses by regularizing attention entropy toward a target value. This differs from sequence-modeling methods like Decision Transformer \citep{chen2021decision}, which learn value-like representations implicitly via return-conditioned prediction rather than Bellman backups. These methods differ in how they represent the value and in the network that computes it, but all of them predict a return from a state--action pair. \oursplain{} works in the opposite direction: it models the action given the state and a return bin, and recovers $Q(s,a)$ with Bayes' rule. Decision Transformer~\citep{chen2021decision} also models actions given a return, but it learns from MC returns-to-go and acts by conditioning on a desired return, whereas \oursplain{} learns from bootstrapped Bellman targets and uses its action model to score candidate actions.

\section{Preliminaries} \label{sec:prelim}

\paragraph{Reinforcement learning.}
We consider a discounted Markov decision process (MDP) specified by the tuple
$\{\mathcal{S}, \mathcal{A}, P, r, \rho_0, \gamma\}$, where $\mathcal{S}$ denotes the
state space, $\mathcal{A}$ the action space, $P(s'\mid s,a)$ the state-transition
distribution, $r:\mathcal{S}\times\mathcal{A}\rightarrow\mathbb{R}$ the one-step reward
function, $\rho_0(s)$ the distribution over initial states, and $\gamma\in[0,1)$ the
discount factor. A policy $\pi(a\mid s)$ induces the discounted return
$G_t=\sum_{k\geq 0}\gamma^{k} r(s_{t+k}, a_{t+k})$, and the goal of RL is to find a
policy that maximizes $\mathbb{E}_{\pi}\left[G_0\right]$. The associated action-value
function is $Q^{\pi}(s,a)=\mathbb{E}_{\pi}\left[G_t \mid s_t=s, a_t=a\right]$. Distributional RL models the conditional distribution of the return $G_t$, rather than its expectation $Q^{\pi}(s,a)$. We use a categorical approximation with support $\mathcal Z=\{z_1,\ldots,z_K\}$ and we refer to each support point $z_i$ as a return bin. For $z\in\mathcal Z$, $p(z\mid s,a)$ denotes the probability mass at $z$.

In this paper, we study how to learn better value functions, asking whether the actions themselves, rather than only the scalar return, can serve as supervision for $Q$, and how this affects downstream control performance. We focus on off-policy RL, where the Q-function estimates the discounted return of the
policy given a state and action, and is learned from a replay buffer $\mathcal{D}$ of
one-step transitions $(s,a,r,s',m)$, with $m\in\{0,1\}$ masking bootstrapping at
terminal transitions. The critic is trained with TD learning:
\begin{equation}
    \mathcal{L} = \mathbb{E}_{(s,a,r,s',m) \sim \mathcal{D}}
    \left[\left(r + \gamma\, m\, \bar Q(s', \tilde{a}') - Q(s, a)\right)^2\right],
\end{equation}
\looseness=-1
where $\bar Q$ is a frozen target copy of the critic $Q$ and $\tilde{a}'\sim\pi(\cdot\mid s')$ is the next
action selected by the RL policy.

\paragraph{Flow matching.}
A conditional flow transforms a simple base variable $x_0\sim\mathcal N(0,I)$ into data $x_1\sim p_{\mathrm{data}}(\cdot\mid c)$. With flow time $\lambda\sim\mathcal U[0,1]$, we use the straight path
\begin{equation}
    x_\lambda=(1-\lambda)x_0+\lambda x_1,
    \qquad u_\lambda=x_1-x_0,
\end{equation}
and train a vector field $v_\theta(c,x_\lambda,\lambda)$ with
\begin{equation}
    \mathcal L_{\mathrm{FM}}(\theta)
    =\mathbb E\!\left[\left\|v_\theta(c,x_\lambda,\lambda)-u_\lambda\right\|_2^2\right].
    \label{eq:fm}
\end{equation}
Sampling integrates $\dot x_\lambda=v_\theta(c,x_\lambda,\lambda)$ from Gaussian noise at $\lambda=0$ to data at $\lambda=1$.

\paragraph{Generative Flow Classifier.}
A class-conditional flow can also classify a fixed observation $x$. For each class $c$, we noise $x$ along the same path and estimate its flow residual
\begin{equation}
    E_\theta(x,c)=\frac{1}{|\mathcal B|}\sum_{(\epsilon,\lambda)\in\mathcal B}
    \frac{1}{D}\left\|v_\theta\!\left(c,(1-\lambda)\epsilon+\lambda x,\lambda\right)-(x-\epsilon)\right\|_2^2.
    \label{eq:flow-class-error}
\end{equation}
This term is a likelihood-like energy and not an exact normalizing-flow density. Reusing the same noise--time pairs $\mathcal B$ for every class gives the proxy, with inverse temperature $\beta>0$:
\begin{equation}
    \widetilde p_\theta(c\mid x)
    =\mathrm{softmax}_c\!\left(-\beta E_\theta(x,c)+\log p(c)\right).
    \label{eq:flow-classifier}
\end{equation}

Given a uniform class prior, the classifier selects the class with the smallest noising error. This is the flow-matching analogue of diffusion classifiers, which compare denoising errors across conditions~\citep{liYourDiffusionModel2023a,clarkTexttoImageDiffusionModels2023b}.

\section{Generative Value Classifier} \label{sec:method}

\begin{figure}[t]
    \centering
    \setlength{\abovecaptionskip}{10pt}
    \makebox[\linewidth][c]{\includegraphics[width=\linewidth]{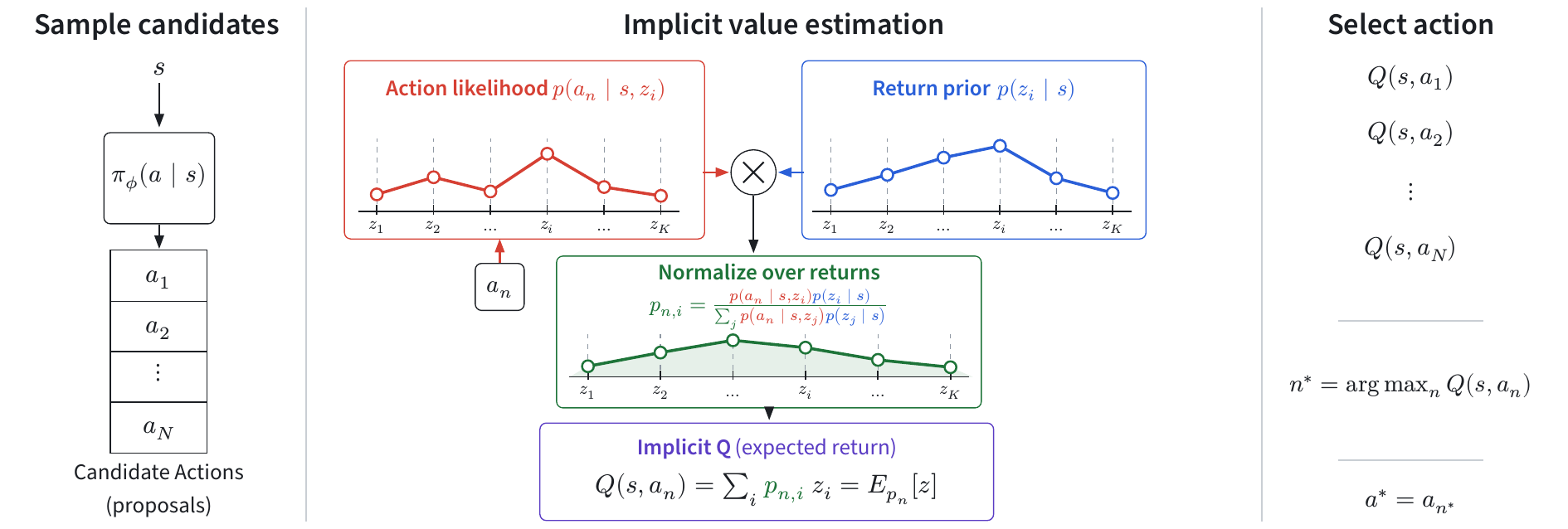}}
    \caption{\textbf{Generative Value Classifier at inference time.} The actor proposes candidate actions (left). For each action, we compute the return-conditioned action likelihood, along with the return prior. Normalizing their product over return bins (center) provides a return distribution whose mean gives $Q(s,a)$, which we use to rank the action candidates (right).}
    \label{fig:gvc-overview}
\end{figure}

\paragraph{Overview.}
\oursplain{} is an implicit generative critic: rather than predicting the return of a state--action pair directly, it models the likelihood of an action under a set of discretized returns and inverts this relationship with Bayes' rule. Conceptually, this changes the question from ``What is the expected return given this state--action pair?'' to ``How likely is this action given this state and return?''

The critic consists of two models: $p_\theta(a\mid s,z)$ assigns a likelihood to a given action under each return bin, and $p_\psi(z\mid s)$ predicts the return distribution from the state alone. The latter is the \textit{return prior}: a distributional state-value function. For each return bin, we multiply the action likelihood by its prior probability, normalize across bins, and take the mean to obtain $Q(s,a)$.

A separate flow-matching actor, trained by behavior cloning on dataset or replay actions, proposes candidate actions (\Cref{ssec:objectives}). The critic scores these candidates, and we execute the one with the highest $Q(s,a)$. \Cref{alg:gvc} summarizes scoring, training, and action selection.

\subsection{From discriminative values to a generative critic}
A conventional distributional critic directly models $p(z\mid s,a)$ over $\mathcal Z$. We instead hold the state fixed and ask the reverse question: how likely is the action $a$ under each return $z$? With $s$ as context, Bayes' rule relates the two directly:
\begin{equation}
    p(z\mid s,a)
    =\frac{p(a\mid s,z)\,p(z\mid s)}{p(a\mid s)}
    \propto p(a\mid s,z)\,p(z\mid s),
    \label{eq:bayes-factorization}
\end{equation}
\looseness=-1
where the denominator $p(a\mid s)$ does not depend on $z$, so normalizing $p(a\mid s,z)\,p(z\mid s)$ over $z$ recovers $p(z\mid s,a)$. The critic therefore needs only two models: the return-conditioned action model $p_\theta(a\mid s,z)$ and the return prior $p_\psi(z\mid s)$. Treating the state as part of the observation instead, $p(s,a\mid z)$, would require a generative model of states; conditioning on $s$ avoids this, and we found it to perform better empirically.

Let $p$ and $Q$ denote the learned return posterior and its mean under $(\theta,\psi)$, with target copies $\bar p,\bar Q$:
\begin{equation}
    p(z_i\mid s,a)
    =\frac{p_\theta(a\mid s,z_i)p_\psi(z_i\mid s)}
           {\sum_j p_\theta(a\mid s,z_j)p_\psi(z_j\mid s)},
    \qquad
    Q(s,a)=\sum_i p(z_i\mid s,a)z_i.
    \label{eq:factorized-return-posterior}
\end{equation}

To evaluate Eq.~\eqref{eq:factorized-return-posterior}, for each bin $z_i$ we take $p_\psi(z_i\mid s)$ from the prior, estimate $p_\theta(a\mid s,z_i)$ from the flow-matching error of $a$ conditioned on $(s,z_i)$, and then normalize over bins. We first introduce this flow-matching surrogate, then the objectives that train both models and align the resulting return.

\begin{figure}[t]
\begingroup
\algrenewcommand{\algorithmicrequire}{\textbf{Given:}}
\algrenewcommand{\alglinenumber}[1]{\scriptsize\color{gray}#1}
\noindent\begin{minipage}{\linewidth}
\fontsize{10}{12}\selectfont
\setlength{\parindent}{0pt}\setlength{\parskip}{0pt}
\hrule height .8pt\vspace{4pt}
\refstepcounter{algorithm}\label{alg:gvc}
\textbf{Algorithm \thealgorithm.} Generative Value Classifier
\vspace{3pt}\hrule height .4pt
\begin{algorithmic}[1]
\Function{Score$_{\theta,\psi}$}{$s,a$}
    \State $\epsilon\sim\mathcal N(0,I_d)$, $\lambda\sim\mathcal U[0,1]$ \Comment{One draw for clarity; shared across bins}
    \State $p_i\gets\operatorname{softmax}_i[\beta_{\mathrm{ret}}\log p_\theta(a\mid s,z_i)+\beta_{\mathrm{prior}}\log p_\psi(z_i\mid s)]$ \Comment{NLL surrogate from Eq.~\eqref{eq:action-flow-error}}
    \State \Return $(p,\sum_ip_i z_i)$ \Comment{$p(z\mid s,a)$ and $Q(s,a)$}
\EndFunction
\Statex
\Function{Train}{$(s,a,r,s',m)\sim\mathcal D$} \Comment{$\mathcal D$: offline dataset or replay buffer}
    \State Sample $a'$ given $s'$. \Comment{e.g., Best-of-$N$ sampling using policy $\phi$}
    \State $(p,Q)\gets{}$\Call{Score$_{\theta,\psi}$}{$s,a$};\quad $(p',Q')\gets{}$\Call{Score$_{\bar\theta,\bar\psi}$}{$s',a'$}
    \State $\widehat p\gets\operatorname{sg}[\Pi_{\mathcal Z}(r+\gamma m Z')]$;\quad $y\gets\operatorname{sg}(\sum_i\widehat p_i z_i)$ \Comment{$\Pr(Z'=z_i)=p'_i$}
    \State $\mathcal L_{\mathrm{flow}}\gets-\!\sum_i\widehat p_i \log p_\theta(a\mid s,z_i)$ \Comment{Critic: Conditional Flow loss}
    \State $\mathcal L_{\mathrm{return}}\gets-\!\sum_i\widehat p_i\log p_\psi(z_i\mid s)$ \Comment{Critic: Categorical Value loss}
    \State $\mathcal L_{\mathrm{anchor}}\gets(Q-y)^2$ \Comment{Critic: Value Anchoring loss}
    \State $\mathcal L_{\mathrm{actor}}\gets-\!\log p_\phi(a\mid s)$ \Comment{Actor: BC Flow loss; Eq.~\eqref{eq:actor-loss}}
    \State Minimize $\mathcal L=\mathcal L_{\mathrm{actor}}+\lambda_F\mathcal L_{\mathrm{flow}}+\lambda_R\mathcal L_{\mathrm{return}}+\lambda_V\mathcal L_{\mathrm{anchor}}$.
    \State Update target copies $(\bar\theta,\bar\psi)$ with EMA.
\EndFunction
\Statex
\Function{Act}{$s$}
    \State Integrate $f_\phi$ given $N$ seeds, $\epsilon\sim\mathcal N(0,I_d)$, to obtain $\mathcal A_N$.
    \State \Return $\arg\max_{a\in\mathcal A_N}Q(s,a)$, with $Q$ from \Call{Score$_{\theta,\psi}$}{$s,a$}. \Comment{Execute the best action}
\EndFunction
\end{algorithmic}
\vspace{2pt}\hrule height .8pt
\end{minipage}\par
\endgroup
\end{figure}

\subsection{Flow likelihood}
\label{ssec:flowlikelihood}
We use the conditional flow-matching error as a surrogate for the negative log-likelihood~\citep{kingma2023understanding,mcallister2025flowmatchingpolicygradients}, writing its negative as $\log p_\theta(a\mid s,z_i)$ below. The vector field $g_\theta(s,x,\lambda,z_i)$ is evaluated for every return bin in a single forward pass: a shared trunk encodes $(s,x,\lambda)$, and each bin has its corresponding learned embedding added to this before a shared head.
\begin{equation}
    -\!\log p_\theta(a\mid s,z_i)
    =\mathbb E_{\epsilon,\lambda}\!\left[
    \frac{1}{d}\left\|g_\theta\!\left(s,(1-\lambda)\epsilon+\lambda a,\lambda,z_i\right)-(a-\epsilon)\right\|_2^2
    \right].
    \label{eq:action-flow-error}
\end{equation}

Substituting this surrogate into Eq.~\eqref{eq:factorized-return-posterior}, with an exponent on each factor, gives
\begin{equation}
    p(z_i\mid s,a)\propto p_\theta(a\mid s,z_i)^{\beta_{\mathrm{ret}}}\,p_\psi(z_i\mid s)^{\beta_{\mathrm{prior}}},
    \label{eq:return-posterior}
\end{equation}
where $\beta_{\mathrm{ret}}$ and $\beta_{\mathrm{prior}}$ are hyperparameters that weight the two factors; both would be $1$ for exact likelihoods. Because the flow error is only a likelihood-like energy, we tune them instead and use a value anchor (\Cref{ssec:objectives}) to fit the mean $Q(s,a)$ to the TD target, since the mean is what ranks actions. \Cref{alg:gvc} shows a single noise--time draw for clarity; when scoring, we average over multiple stratified flow timesteps with antithetic noise, shared across return bins and action candidates to reduce variance~\citep{liYourDiffusionModel2023a,chenDiffusionModelsAre2024}.

\subsection{Objectives}
\label{ssec:objectives}
\paragraph{Critic objectives.}
For each transition, we choose a next action $a'$, for example via best-of-$N$ sampling. Let $\widehat p$ be the distribution of $r+\gamma Z'$ projected onto $\mathcal Z$, where $Z'\sim\bar p(\cdot\mid s',a')$. The projection clips shifted atoms to $[z_1,z_K]$ (\Cref{app:gvc-hparams}).

We use this distribution as a soft target, weighting each bin by $\widehat p_i$ to supervise both the return-conditioned generative loss and the return prior:
\begin{align}
    \mathcal L_{\mathrm{flow}}
    &=\mathbb E_{\mathcal D}\!\left[-\!\sum_i\widehat p_i \log p_\theta(a\mid s,z_i)\right], \\
    \mathcal L_{\mathrm{return}}
    &=\mathbb E_{\mathcal D}\!\left[-\!\sum_i\widehat p_i
      \log p_\psi(z_i\mid s)\right],
    \label{eq:base-critic-losses}
\end{align}
where $\mathcal L_{\mathrm{flow}}$ is just the flow-matching loss of Eq.~\eqref{eq:action-flow-error} on the transition's action $a$ (a fresh noise--time sample per transition), with each return bin weighted by its TD-target probability $\widehat p_i$, and $\mathcal L_{\mathrm{return}}$ fits the prior to the same target. The flow error is only a surrogate, with unknown scale and bin-dependent offsets, so these losses alone do not ensure that $Q(s,a)$ matches the target mean $y=\sum_i\widehat p_i z_i$. To penalize the value error directly, we add a squared-error value anchor:
\begin{align}
    \mathcal L_{\mathrm{anchor}}
    &=\mathbb E_{\mathcal D}\!\left[
      \left(Q(s,a)-\mathrm{sg}(y)\right)^2\right].
    \label{eq:value-anchors}
\end{align}
The value anchor constrains the return which is used to rank actions.

\paragraph{Actor Objective.} \oursplain{} is independent of the policy extraction mechanism and can work with both implicit and explicit extraction methods. For all experiments, except the ablation in \Cref{app:actor-distillation}, we use best-of-$N$ sampling for implicit extraction alongside a flow-matching behavior policy trained on replay actions, both to act and to select $a'$ for the TD target; \Cref{app:gvc-hparams} lists $N$ for each setting. %

The actor is a behavior-cloning flow with vector field $f_\phi$. Its loss is $-\mathbb E_{\mathcal D}\log p_\phi(a\mid s)$, using the flow-matching surrogate
\begin{equation}
    -\!\log p_\phi(a\mid s)
    =\mathbb E_{\epsilon,\lambda}\!\left[
      \frac{1}{d}\left\|f_\phi\!\left(s,(1-\lambda)\epsilon+\lambda a,\lambda\right)-(a-\epsilon)\right\|_2^2\right].
    \label{eq:actor-loss}
\end{equation}
\paragraph{Complete Objective.}
The complete update minimizes
\begin{equation}
    \mathcal L
    =\mathcal L_{\mathrm{actor}}+\lambda_F\mathcal L_{\mathrm{flow}}
     +\lambda_R\mathcal L_{\mathrm{return}}
     +\lambda_V\mathcal L_{\mathrm{anchor}}.
    \label{eq:total-loss}
\end{equation}
At evaluation, we sample $N$ distinct actions from the actor, score them using Eq.~\eqref{eq:return-posterior}, execute the action with the largest $Q(s,a)$, and replan at the next environment step.

\section{Experiments} \label{sec:exp}

The goal of our experiments is to answer the following key questions:

\begin{enumerate}[start=1,label={(\bfseries Q\arabic*)}]
    \item Does replacing a discriminative critic with \oursplain{} improve online RL when the actor and training recipe are held fixed?
    \item Does the gain hold in offline RL and during offline-to-online finetuning, including against distributional critics?
    \item Which parts of the \oursplain{} objective are responsible for the gains?
\end{enumerate}

\textbf{Environments.}
We evaluate \oursplain{} in two settings.
For online RL, we use four robotic manipulation tasks---Lift, Can, Square, and Tool Hang---from RoboMimic~\citep{mandlekar2021matterslearningofflinehuman}, where a 7-DoF robot arm must perform diverse manipulation behaviors under sparse task-completion rewards.
Lift requires the robot to grasp and lift up an object; Can requires grasping a cylinder and placing it at a specified location; Square involves precisely inserting a square nut onto a peg; Tool Hang is a long-horizon, multi-stage assembly task in which the robot must insert a hook into a base to make a frame and then hang a wrench from it.
For offline and offline-to-online RL, we evaluate on OGBench~\citep{park2025ogbench}, which provides offline datasets paired with online finetuning for long-horizon, sparse-reward control. Offline, we use the \texttt{cube}, \texttt{puzzle}, and sparse-reward \texttt{scene} domains listed in \Cref{tab:offline-eval-full}; for offline-to-online finetuning we use the six tasks from Value Flows~\citep{dong2026valueflows}.

\begin{figure*}[t]
\centering
\includegraphics[width=\textwidth]{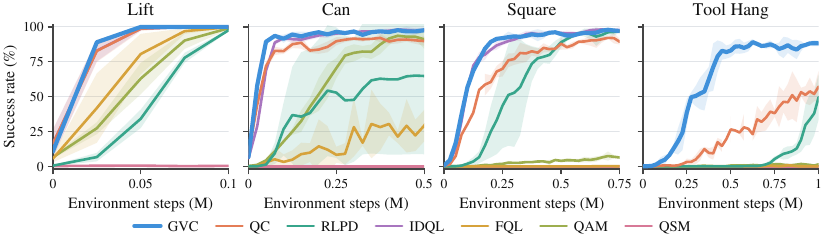}
\caption{\textbf{Online RL on RoboMimic.} Success rate versus environment steps (in millions) on Lift, Can, Square, and Tool Hang, from left to right. We compare \oursplain{} with QC, RLPD, IDQL, FQL, QAM, and QSM. Shaded regions show one standard deviation across seeds.}
\label{fig:robomimic-online}
\end{figure*}

\textbf{Baselines.}
We organize the baselines based on what they change relative to \oursplain{}. BFN~\citep{li2025actionchunking} keeps the flow-matching actor and the best-of-$N$ selection of \oursplain{} and swaps the critic for a regression head, so the two differ only in the critic. IDQL~\citep{hansenestruch2023idqlimplicitqlearningactorcritic} and its flow-policy variant IFQL~\citep{park2025flowqlearning} also select actions by best-of-$N$ from a behavior-cloned generative policy but fit their critic by expectile regression, without querying the policy. C51~\citep{bellemare2017distributional}, IQN~\citep{dabney18implicit}, CODAC~\citep{ma2021conservative}, and Value Flows~\citep{dong2026valueflows} keep a discriminative head but model the full return distribution, with categorical, quantile, conservative quantile, and flow-matching parameterizations respectively; they test whether distributional targets alone explain our gains. FQL~\citep{park2025flowqlearning}, QSM~\citep{pmlr-v235-psenka24a}, and QAM~\citep{li2026q} instead keep a standard critic and train an explicit flow or diffusion policy against it, by distillation with a BC penalty, score matching, or adjoint matching, while QC~\citep{li2025actionchunking} runs TD learning over action chunks to shorten the effective horizon. Finally, we include RLPD~\citep{ball2023efficient}, IQL~\citep{kostrikov2022offline}, and, offline only, BC, ReBRAC~\citep{tarasov2023revisiting}, and FBRAC~\citep{wu2019behavior,park2025flowqlearning} as standard references; offline baseline numbers are taken from Value Flows or reproduced directly where unavailable (\Cref{app:baseline-tuning}). RLPD and FQL are also run offline-to-online, initialized from the offline dataset and finetuned online.

\textbf{Implementation details.}
To fairly compare \oursplain{} to prior methods, we follow the same training recipe as QC~\citep{li2025actionchunking} for the RoboMimic online experiments and Value Flows~\citep{dong2026valueflows} for the OGBench offline and offline-to-online experiments. We utilize the same BFN actor architecture, only modifying the critic architecture to implement our method. For RoboMimic and OGBench \texttt{cube} environments, we predict an advantage instead of the absolute Q-value to reduce binning error. \Cref{app:gvc-hparams} lists the \oursplain{} hyperparameters for each setting, \Cref{app:protocol} the training and evaluation protocol, and \Cref{app:baseline-tuning} the baseline tuning.

\subsection{Online RL}

\Cref{fig:robomimic-online} shows learning curves on the four RoboMimic tasks. Lift, Can, and Square are eventually solved by \oursplain{}, QC, and IDQL, although \oursplain{} demonstrates small sample-efficiency gains over all curves. Tool Hang is where the critic methods diverge, with \oursplain{} reaching roughly 85\% success within 0.5M steps whereas QC, whose recipe \oursplain{} shares apart from the critic, and RLPD only reach about 50\% at 1M steps. Meanwhile, IDQL, which uses the same best-of-$N$ selection but a regression critic, does not learn the task. %

\begin{table}[t]
\caption{\footnotesize \textbf{Offline evaluation on OGBench.} Success rate (\%) per task. We bold values within $95\%$ of the best performance for each task.}
\label{tab:offline-eval-full}
\centering
\begingroup
\newcommand{\ogunc}[1]{\,\text{\scalebox{.80}{\color{black!65}$\pm #1$}}}

\setlength{\tabcolsep}{4pt}
\renewcommand{\arraystretch}{1.04}
\resizebox{\linewidth}{!}{%
\begin{tabular*}{1.63\linewidth}{@{\extracolsep{\fill}}llcccccccccccc}
\toprule
\multicolumn{2}{l}{} & \multicolumn{3}{c}{Gaussian Policies} & \multicolumn{9}{c}{Flow Policies} \\
\cmidrule(lr){3-5} \cmidrule(lr){6-14}
Environment & Task & BC & IQL & ReBRAC & FBRAC & IFQL & FQL & C51 & IQN & CODAC & Value Flows & BFN & GVC \\
\midrule
\multirow{6}{*}{\texttt{scene-sparse}} & \texttt{task1} & $19\ogunc{6}$ & $0\ogunc{0}$ & $9\ogunc{8}$ & $\mathbf{100}\ogunc{1}$ & $\mathbf{96}\ogunc{3}$ & $65\ogunc{16}$ & $19\ogunc{6}$ & $\mathbf{99}\ogunc{2}$ & $\mathbf{100}\ogunc{1}$ & $\mathbf{100}\ogunc{1}$ & $\mathbf{100}\ogunc{1}$ & $\mathbf{99}\ogunc{1}$ \\
 & \texttt{task2} & $1\ogunc{1}$ & $0\ogunc{0}$ & $66\ogunc{19}$ & $86\ogunc{10}$ & $83\ogunc{5}$ & $52\ogunc{12}$ & $2\ogunc{2}$ & $\mathbf{98}\ogunc{2}$ & $\mathbf{98}\ogunc{3}$ & $\mathbf{99}\ogunc{1}$ & $\mathbf{100}\ogunc{1}$ & $\mathbf{99}\ogunc{2}$ \\
 & \texttt{task3} & $1\ogunc{1}$ & $0\ogunc{0}$ & $0\ogunc{0}$ & $\mathbf{94}\ogunc{8}$ & $56\ogunc{8}$ & $60\ogunc{2}$ & $2\ogunc{2}$ & $93\ogunc{6}$ & $72\ogunc{38}$ & $89\ogunc{5}$ & $\mathbf{94}\ogunc{6}$ & $\mathbf{99}\ogunc{2}$ \\
 & \texttt{task4} & $2\ogunc{2}$ & $0\ogunc{0}$ & $0\ogunc{0}$ & $77\ogunc{34}$ & $72\ogunc{12}$ & $69\ogunc{16}$ & $4\ogunc{4}$ & $82\ogunc{15}$ & $76\ogunc{30}$ & $92\ogunc{10}$ & $90\ogunc{4}$ & $\mathbf{98}\ogunc{3}$ \\
 & \texttt{task5} & $0\ogunc{0}$ & $0\ogunc{0}$ & $0\ogunc{0}$ & $75\ogunc{19}$ & $52\ogunc{8}$ & $19\ogunc{5}$ & $1\ogunc{2}$ & $57\ogunc{30}$ & $66\ogunc{20}$ & $91\ogunc{7}$ & $45\ogunc{14}$ & $\mathbf{97}\ogunc{3}$ \\
\cmidrule(l){3-14}
 & \texttt{avg.} & $5$ & $0$ & $15$ & $86$ & $72$ & $53$ & $6$ & $86$ & $82$ & $\mathbf{94}$ & $86$ & $\mathbf{98}$ \\
\midrule
\multirow{6}{*}{\texttt{puzzle3x3-sparse}} & \texttt{task1} & $5\ogunc{2}$ & $1\ogunc{1}$ & $\mathbf{98}\ogunc{2}$ & $75\ogunc{19}$ & $\mathbf{99}\ogunc{1}$ & $\mathbf{100}\ogunc{0}$ & $6\ogunc{3}$ & $37\ogunc{29}$ & $5\ogunc{4}$ & $78\ogunc{15}$ & $\mathbf{99}\ogunc{1}$ & $\mathbf{100}\ogunc{0}$ \\
 & \texttt{task2} & $1\ogunc{1}$ & $0\ogunc{0}$ & $66\ogunc{15}$ & $35\ogunc{37}$ & $\mathbf{99}\ogunc{2}$ & $\mathbf{100}\ogunc{0}$ & $3\ogunc{3}$ & $13\ogunc{18}$ & $1\ogunc{1}$ & $5\ogunc{7}$ & $\mathbf{97}\ogunc{5}$ & $\mathbf{100}\ogunc{0}$ \\
 & \texttt{task3} & $1\ogunc{1}$ & $0\ogunc{0}$ & $25\ogunc{16}$ & $21\ogunc{22}$ & $\mathbf{98}\ogunc{2}$ & $\mathbf{97}\ogunc{4}$ & $0\ogunc{1}$ & $41\ogunc{48}$ & $1\ogunc{1}$ & $10\ogunc{5}$ & $\mathbf{95}\ogunc{5}$ & $\mathbf{97}\ogunc{4}$ \\
 & \texttt{task4} & $1\ogunc{1}$ & $0\ogunc{0}$ & $55\ogunc{13}$ & $41\ogunc{34}$ & $\mathbf{98}\ogunc{2}$ & $\mathbf{98}\ogunc{2}$ & $0\ogunc{1}$ & $2\ogunc{2}$ & $1\ogunc{2}$ & $8\ogunc{6}$ & $\mathbf{98}\ogunc{3}$ & $\mathbf{98}\ogunc{1}$ \\
 & \texttt{task5} & $1\ogunc{0}$ & $0\ogunc{0}$ & $68\ogunc{18}$ & $43\ogunc{40}$ & $\mathbf{97}\ogunc{4}$ & $\mathbf{100}\ogunc{0}$ & $1\ogunc{1}$ & $1\ogunc{2}$ & $0\ogunc{1}$ & $4\ogunc{4}$ & $\mathbf{96}\ogunc{2}$ & $\mathbf{96}\ogunc{4}$ \\
\cmidrule(l){3-14}
 & \texttt{avg.} & $2$ & $0$ & $62$ & $43$ & $\mathbf{98}$ & $\mathbf{99}$ & $2$ & $19$ & $2$ & $21$ & $\mathbf{97}$ & $\mathbf{98}$ \\
\midrule
\multirow{6}{*}{\texttt{puzzle4x4}} & \texttt{task1} & $1\ogunc{1}$ & $12\ogunc{2}$ & $26\ogunc{4}$ & $32\ogunc{9}$ & $49\ogunc{9}$ & $34\ogunc{8}$ & $0\ogunc{0}$ & $41\ogunc{2}$ & $37\ogunc{32}$ & $36\ogunc{3}$ & $62\ogunc{10}$ & $\mathbf{76}\ogunc{4}$ \\
 & \texttt{task2} & $0\ogunc{0}$ & $7\ogunc{4}$ & $12\ogunc{4}$ & $5\ogunc{3}$ & $4\ogunc{4}$ & $16\ogunc{5}$ & $0\ogunc{0}$ & $12\ogunc{4}$ & $10\ogunc{10}$ & $\mathbf{27}\ogunc{5}$ & $20\ogunc{7}$ & $19\ogunc{12}$ \\
 & \texttt{task3} & $0\ogunc{0}$ & $9\ogunc{3}$ & $15\ogunc{3}$ & $20\ogunc{10}$ & $50\ogunc{14}$ & $18\ogunc{5}$ & $0\ogunc{0}$ & $45\ogunc{7}$ & $33\ogunc{29}$ & $30\ogunc{4}$ & $41\ogunc{8}$ & $\mathbf{63}\ogunc{12}$ \\
 & \texttt{task4} & $0\ogunc{0}$ & $5\ogunc{2}$ & $10\ogunc{3}$ & $5\ogunc{1}$ & $21\ogunc{11}$ & $11\ogunc{3}$ & $0\ogunc{0}$ & $23\ogunc{2}$ & $12\ogunc{10}$ & $\mathbf{28}\ogunc{5}$ & $21\ogunc{6}$ & $\mathbf{28}\ogunc{11}$ \\
 & \texttt{task5} & $0\ogunc{0}$ & $4\ogunc{1}$ & $7\ogunc{3}$ & $2\ogunc{2}$ & $2\ogunc{2}$ & $7\ogunc{3}$ & $0\ogunc{0}$ & $16\ogunc{6}$ & $10\ogunc{8}$ & $13\ogunc{2}$ & $14\ogunc{6}$ & $\mathbf{19}\ogunc{9}$ \\
\cmidrule(l){3-14}
 & \texttt{avg.} & $0$ & $7$ & $14$ & $13$ & $25$ & $17$ & $0$ & $27$ & $20$ & $27$ & $32$ & $\mathbf{41}$ \\
\midrule
\multirow{6}{*}{\texttt{puzzle4x5}} & \texttt{task1} & $0\ogunc{0}$ & $14\ogunc{7}$ & $0\ogunc{1}$ & $32\ogunc{30}$ & $86\ogunc{8}$ & $53\ogunc{19}$ & $0\ogunc{0}$ & $\mathbf{90}\ogunc{5}$ & $51\ogunc{10}$ & $67\ogunc{4}$ & $83\ogunc{4}$ & $\mathbf{92}\ogunc{4}$ \\
 & \texttt{task2} & $0\ogunc{0}$ & $0\ogunc{0}$ & $0\ogunc{0}$ & $0\ogunc{0}$ & $0\ogunc{0}$ & $0\ogunc{0}$ & $0\ogunc{0}$ & $0\ogunc{1}$ & $\mathbf{1}\ogunc{2}$ & $0\ogunc{1}$ & $1\ogunc{1}$ & $0\ogunc{0}$ \\
 & \texttt{task3} & $0\ogunc{0}$ & $0\ogunc{0}$ & $0\ogunc{0}$ & $0\ogunc{0}$ & $0\ogunc{0}$ & $0\ogunc{0}$ & $0\ogunc{0}$ & $0\ogunc{0}$ & $\mathbf{0}\ogunc{1}$ & $\mathbf{0}\ogunc{1}$ & $0\ogunc{0}$ & $0\ogunc{0}$ \\
 & \texttt{task4} & $\mathbf{0}\ogunc{0}$ & $\mathbf{0}\ogunc{0}$ & $\mathbf{0}\ogunc{0}$ & $\mathbf{0}\ogunc{0}$ & $\mathbf{0}\ogunc{0}$ & $\mathbf{0}\ogunc{0}$ & $\mathbf{0}\ogunc{0}$ & $\mathbf{0}\ogunc{0}$ & $\mathbf{0}\ogunc{0}$ & $\mathbf{0}\ogunc{0}$ & $\mathbf{0}$ & $\mathbf{0}\ogunc{0}$ \\
 & \texttt{task5} & $\mathbf{0}\ogunc{0}$ & $\mathbf{0}\ogunc{0}$ & $\mathbf{0}\ogunc{0}$ & $\mathbf{0}\ogunc{0}$ & $\mathbf{0}\ogunc{0}$ & $\mathbf{0}\ogunc{0}$ & $\mathbf{0}\ogunc{0}$ & $\mathbf{0}\ogunc{0}$ & $\mathbf{0}\ogunc{0}$ & $\mathbf{0}\ogunc{0}$ & $\mathbf{0}\ogunc{0}$ & $\mathbf{0}\ogunc{0}$ \\
\cmidrule(l){3-14}
 & \texttt{avg.} & $0$ & $3$ & $0$ & $6$ & $17$ & $11$ & $0$ & $\mathbf{18}$ & $10$ & $13$ & $17$ & $\mathbf{18}$ \\
\midrule
\multirow{6}{*}{\texttt{puzzle4x6}} & \texttt{task1} & $0\ogunc{0}$ & $2\ogunc{3}$ & $8\ogunc{7}$ & $16\ogunc{20}$ & $0\ogunc{0}$ & $22\ogunc{5}$ & $0\ogunc{0}$ & $30\ogunc{14}$ & $10\ogunc{11}$ & $22\ogunc{25}$ & $22\ogunc{8}$ & $\mathbf{48}\ogunc{16}$ \\
 & \texttt{task2} & $0\ogunc{0}$ & $0\ogunc{1}$ & $0\ogunc{1}$ & $0\ogunc{1}$ & $0\ogunc{0}$ & $0\ogunc{1}$ & $0\ogunc{0}$ & $0\ogunc{1}$ & $\mathbf{2}\ogunc{2}$ & $0\ogunc{1}$ & $1\ogunc{1}$ & $0\ogunc{0}$ \\
 & \texttt{task3} & $\mathbf{0}\ogunc{0}$ & $\mathbf{0}\ogunc{0}$ & $\mathbf{0}\ogunc{0}$ & $\mathbf{0}\ogunc{0}$ & $\mathbf{0}\ogunc{0}$ & $\mathbf{0}\ogunc{0}$ & $\mathbf{0}\ogunc{0}$ & $\mathbf{0}\ogunc{0}$ & $\mathbf{0}\ogunc{0}$ & $\mathbf{0}\ogunc{0}$ & $\mathbf{0}\ogunc{0}$ & $\mathbf{0}\ogunc{0}$ \\
 & \texttt{task4} & $\mathbf{0}\ogunc{0}$ & $\mathbf{0}\ogunc{0}$ & $\mathbf{0}\ogunc{0}$ & $\mathbf{0}\ogunc{0}$ & $\mathbf{0}\ogunc{0}$ & $\mathbf{0}\ogunc{0}$ & $\mathbf{0}\ogunc{0}$ & $\mathbf{0}\ogunc{0}$ & $\mathbf{0}\ogunc{0}$ & $\mathbf{0}\ogunc{0}$ & $\mathbf{0}\ogunc{0}$ & $\mathbf{0}\ogunc{0}$ \\
 & \texttt{task5} & $\mathbf{0}\ogunc{0}$ & $\mathbf{0}\ogunc{0}$ & $\mathbf{0}\ogunc{0}$ & $\mathbf{0}\ogunc{0}$ & $\mathbf{0}\ogunc{0}$ & $\mathbf{0}\ogunc{0}$ & $\mathbf{0}\ogunc{0}$ & $\mathbf{0}\ogunc{0}$ & $\mathbf{0}\ogunc{0}$ & $\mathbf{0}\ogunc{0}$ & $\mathbf{0}$ & $\mathbf{0}\ogunc{0}$ \\
\cmidrule(l){3-14}
 & \texttt{avg.} & $0$ & $0$ & $2$ & $3$ & $0$ & $4$ & $0$ & $6$ & $2$ & $4$ & $5$ & $\mathbf{10}$ \\
\midrule
\multirow{6}{*}{\texttt{cube-double}} & \texttt{task1} & $8\ogunc{3}$ & $27\ogunc{5}$ & $45\ogunc{6}$ & $47\ogunc{11}$ & $35\ogunc{9}$ & $61\ogunc{9}$ & $9\ogunc{0}$ & $70\ogunc{14}$ & $80\ogunc{11}$ & $\mathbf{97}\ogunc{1}$ & $73\ogunc{4}$ & $88\ogunc{21}$ \\
 & \texttt{task2} & $0\ogunc{0}$ & $1\ogunc{1}$ & $7\ogunc{3}$ & $22\ogunc{12}$ & $9\ogunc{5}$ & $36\ogunc{6}$ & $0\ogunc{0}$ & $24\ogunc{9}$ & $63\ogunc{4}$ & $\mathbf{76}\ogunc{7}$ & $\mathbf{78}\ogunc{4}$ & $71\ogunc{12}$ \\
 & \texttt{task3} & $0\ogunc{0}$ & $0\ogunc{0}$ & $4\ogunc{1}$ & $4\ogunc{2}$ & $8\ogunc{5}$ & $22\ogunc{5}$ & $0\ogunc{0}$ & $25\ogunc{6}$ & $66\ogunc{9}$ & $73\ogunc{4}$ & $\mathbf{77}\ogunc{7}$ & $\mathbf{74}\ogunc{26}$ \\
 & \texttt{task4} & $0\ogunc{0}$ & $0\ogunc{0}$ & $1\ogunc{1}$ & $0\ogunc{1}$ & $1\ogunc{1}$ & $5\ogunc{2}$ & $0\ogunc{0}$ & $10\ogunc{1}$ & $13\ogunc{2}$ & $\mathbf{30}\ogunc{5}$ & $23\ogunc{6}$ & $19\ogunc{6}$ \\
 & \texttt{task5} & $0\ogunc{0}$ & $4\ogunc{3}$ & $4\ogunc{2}$ & $2\ogunc{2}$ & $17\ogunc{6}$ & $19\ogunc{10}$ & $0\ogunc{0}$ & $\mathbf{81}\ogunc{8}$ & $\mathbf{82}\ogunc{4}$ & $69\ogunc{5}$ & $\mathbf{79}\ogunc{6}$ & $75\ogunc{19}$ \\
\cmidrule(l){3-14}
 & \texttt{avg.} & $2$ & $6$ & $12$ & $15$ & $14$ & $29$ & $2$ & $42$ & $61$ & $\mathbf{69}$ & $\mathbf{66}$ & $65$ \\
\midrule
\multirow{6}{*}{\texttt{cube-triple}} & \texttt{task1} & $1\ogunc{1}$ & $4\ogunc{4}$ & $1\ogunc{2}$ & $0\ogunc{0}$ & $2\ogunc{2}$ & $20\ogunc{6}$ & $1\ogunc{0}$ & $29\ogunc{2}$ & $9\ogunc{5}$ & $\mathbf{59}\ogunc{12}$ & $28\ogunc{13}$ & $49\ogunc{30}$ \\
 & \texttt{task2} & $0\ogunc{0}$ & $0\ogunc{0}$ & $0\ogunc{0}$ & $0\ogunc{0}$ & $0\ogunc{0}$ & $\mathbf{1}\ogunc{2}$ & $0\ogunc{0}$ & $0\ogunc{0}$ & $\mathbf{1}\ogunc{0}$ & $0\ogunc{0}$ & $1\ogunc{1}$ & $1\ogunc{1}$ \\
 & \texttt{task3} & $0\ogunc{0}$ & $0\ogunc{0}$ & $0\ogunc{0}$ & $0\ogunc{0}$ & $0\ogunc{0}$ & $0\ogunc{0}$ & $0\ogunc{0}$ & $1\ogunc{0}$ & $0\ogunc{0}$ & $\mathbf{7}\ogunc{3}$ & $2\ogunc{0}$ & $2\ogunc{3}$ \\
 & \texttt{task4} & $\mathbf{0}\ogunc{0}$ & $\mathbf{0}\ogunc{0}$ & $\mathbf{0}\ogunc{0}$ & $\mathbf{0}\ogunc{0}$ & $\mathbf{0}\ogunc{0}$ & $\mathbf{0}\ogunc{0}$ & $\mathbf{0}\ogunc{0}$ & $\mathbf{0}\ogunc{0}$ & $\mathbf{0}\ogunc{0}$ & $\mathbf{0}\ogunc{0}$ & $\mathbf{0}\ogunc{0}$ & $\mathbf{0}\ogunc{0}$ \\
 & \texttt{task5} & $0\ogunc{0}$ & $1\ogunc{1}$ & $0\ogunc{0}$ & $0\ogunc{0}$ & $0\ogunc{0}$ & $0\ogunc{0}$ & $0\ogunc{0}$ & $0\ogunc{0}$ & $0\ogunc{0}$ & $\mathbf{2}\ogunc{1}$ & $0\ogunc{0}$ & $0\ogunc{0}$ \\
\cmidrule(l){3-14}
 & \texttt{avg.} & $0$ & $1$ & $0$ & $0$ & $0$ & $4$ & $0$ & $6$ & $2$ & $\mathbf{14}$ & $6$ & $10$ \\
\bottomrule
\end{tabular*}%
}
\endgroup
\end{table}

\subsection{Offline RL}

\looseness=-1
\Cref{tab:offline-eval-full} reports per-task success on 35 tasks across seven OGBench domains. On the two largest puzzles, \texttt{puzzle-4x5} and \texttt{puzzle-4x6}, \oursplain{} reaches 92\% and 48\% on the first task, compared to BFN---the closest comparison, differing from \oursplain{} only in the critic---which achieves 83\% and 22\%, respectively. The two are nearly tied on \texttt{cube-double} (65\%-66\%) and are both near ceiling on the sparse \texttt{puzzle-3x3} tasks; \oursplain{} leads by 12 points on the sparse \texttt{scene} tasks (98\% versus 86\%), mostly on \texttt{task5} where BFN reaches 45\%, and by 4 on \texttt{cube-triple} (10\% versus 6\%). IFQL, which also selects by best-of-$N$ but trains its critic by expectile regression, is further behind: \oursplain{} is ahead on six of the seven domains. \oursplain{} also improves over the leading distributional baseline, Value Flows, on \texttt{puzzle-4x4}, \texttt{puzzle-4x5}, and \texttt{puzzle-4x6} by an average of 8 points, while trailing slightly on \texttt{cube-double} and \texttt{cube-triple} by an average of 4 points.

\begin{table}[t]
\caption{\footnotesize \textbf{Offline-to-online evaluations on OGBench tasks.} Success rate (\%) on 6 OGBench tasks after offline training $\rightarrow$ after online finetuning. Mean $\pm$ standard deviation over 8 seeds, with values within $95\%$ of the best final performance for each task in bold.}
\label{tab:offline2online-eval}
\centering
\begingroup
\newcommand{\oounc}[1]{\,\text{\scalebox{.80}{\color{black!65}$\pm #1$}}}
\small
\setlength{\tabcolsep}{3pt}
\renewcommand{\arraystretch}{1.10}
\resizebox{\linewidth}{!}{%
\begin{tabular}{@{}lcccccccc@{}}
\toprule
Task & IQN & IFQL & FQL & RLPD & IQL & Value Flows & BFN & GVC \\
\midrule
\texttt{antmaze-large} & \makebox[3.15em][r]{$55\oounc{5}$}\,$\rightarrow$\,\makebox[3.15em][l]{$91\oounc{1}$} & \makebox[3.15em][r]{$11\oounc{3}$}\,$\rightarrow$\,\makebox[3.15em][l]{$3\oounc{2}$} & \makebox[3.15em][r]{$\mathbf{87}\oounc{7}$}\,$\rightarrow$\,\makebox[3.15em][l]{$\mathbf{100}\oounc{1}$} & \makebox[3.15em][r]{$0\oounc{0}$}\,$\rightarrow$\,\makebox[3.15em][l]{$\mathbf{99}\oounc{2}$} & \makebox[3.15em][r]{$41\oounc{5}$}\,$\rightarrow$\,\makebox[3.15em][l]{$45\oounc{5}$} & \makebox[3.15em][r]{$\mathbf{72}\oounc{4}$}\,$\rightarrow$\,\makebox[3.15em][l]{$\mathbf{96}\oounc{1}$} & \makebox[3.15em][r]{$\mathbf{66}\oounc{19}$}\,$\rightarrow$\,\makebox[3.15em][l]{$\mathbf{99}\oounc{1}$} & \makebox[3.15em][r]{$\mathbf{77}\oounc{9}$}\,$\rightarrow$\,\makebox[3.15em][l]{$\mathbf{97}\oounc{2}$} \\
\texttt{humanoidmaze-medium} & \makebox[3.15em][r]{$23\oounc{3}$}\,$\rightarrow$\,\makebox[3.15em][l]{$14\oounc{2}$} & \makebox[3.15em][r]{$56\oounc{12}$}\,$\rightarrow$\,\makebox[3.15em][l]{$82\oounc{7}$} & \makebox[3.15em][r]{$12\oounc{7}$}\,$\rightarrow$\,\makebox[3.15em][l]{$22\oounc{12}$} & \makebox[3.15em][r]{$0\oounc{0}$}\,$\rightarrow$\,\makebox[3.15em][l]{$8\oounc{3}$} & \makebox[3.15em][r]{$21\oounc{5}$}\,$\rightarrow$\,\makebox[3.15em][l]{$16\oounc{3}$} & \makebox[3.15em][r]{$29\oounc{6}$}\,$\rightarrow$\,\makebox[3.15em][l]{$86\oounc{3}$} & \makebox[3.15em][r]{$32\oounc{8}$}\,$\rightarrow$\,\makebox[3.15em][l]{$89\oounc{4}$} & \makebox[3.15em][r]{$\mathbf{94}\oounc{2}$}\,$\rightarrow$\,\makebox[3.15em][l]{$\mathbf{99}\oounc{1}$} \\
\texttt{cube-double} & \makebox[3.15em][r]{$29\oounc{4}$}\,$\rightarrow$\,\makebox[3.15em][l]{$42\oounc{7}$} & \makebox[3.15em][r]{$12\oounc{3}$}\,$\rightarrow$\,\makebox[3.15em][l]{$41\oounc{2}$} & \makebox[3.15em][r]{$40\oounc{11}$}\,$\rightarrow$\,\makebox[3.15em][l]{$92\oounc{3}$} & \makebox[3.15em][r]{$0\oounc{0}$}\,$\rightarrow$\,\makebox[3.15em][l]{$0\oounc{0}$} & \makebox[3.15em][r]{$0\oounc{0}$}\,$\rightarrow$\,\makebox[3.15em][l]{$0\oounc{0}$} & \makebox[3.15em][r]{$65\oounc{7}$}\,$\rightarrow$\,\makebox[3.15em][l]{$79\oounc{6}$} & \makebox[3.15em][r]{$79\oounc{7}$}\,$\rightarrow$\,\makebox[3.15em][l]{$90\oounc{7}$} & \makebox[3.15em][r]{$\mathbf{64}\oounc{21}$}\,$\rightarrow$\,\makebox[3.15em][l]{$\mathbf{98}\oounc{1}$} \\
\texttt{cube-triple} & \makebox[3.15em][r]{$25\oounc{5}$}\,$\rightarrow$\,\makebox[3.15em][l]{$14\oounc{6}$} & \makebox[3.15em][r]{$2\oounc{1}$}\,$\rightarrow$\,\makebox[3.15em][l]{$7\oounc{5}$} & \makebox[3.15em][r]{$4\oounc{1}$}\,$\rightarrow$\,\makebox[3.15em][l]{$83\oounc{12}$} & \makebox[3.15em][r]{$0\oounc{0}$}\,$\rightarrow$\,\makebox[3.15em][l]{$0\oounc{0}$} & \makebox[3.15em][r]{$3\oounc{1}$}\,$\rightarrow$\,\makebox[3.15em][l]{$0\oounc{0}$} & \makebox[3.15em][r]{$29\oounc{8}$}\,$\rightarrow$\,\makebox[3.15em][l]{$70\oounc{7}$} & \makebox[3.15em][r]{$21\oounc{10}$}\,$\rightarrow$\,\makebox[3.15em][l]{$90\oounc{17}$} & \makebox[3.15em][r]{$\mathbf{46}\oounc{21}$}\,$\rightarrow$\,\makebox[3.15em][l]{$\mathbf{100}\oounc{0}$} \\
\texttt{puzzle-4x4} & \makebox[3.15em][r]{$22\oounc{2}$}\,$\rightarrow$\,\makebox[3.15em][l]{$6\oounc{1}$} & \makebox[3.15em][r]{$23\oounc{2}$}\,$\rightarrow$\,\makebox[3.15em][l]{$19\oounc{12}$} & \makebox[3.15em][r]{$8\oounc{3}$}\,$\rightarrow$\,\makebox[3.15em][l]{$38\oounc{52}$} & \makebox[3.15em][r]{$\mathbf{0}\oounc{0}$}\,$\rightarrow$\,\makebox[3.15em][l]{$\mathbf{100}\oounc{0}$} & \makebox[3.15em][r]{$5\oounc{1}$}\,$\rightarrow$\,\makebox[3.15em][l]{$0\oounc{0}$} & \makebox[3.15em][r]{$14\oounc{3}$}\,$\rightarrow$\,\makebox[3.15em][l]{$51\oounc{12}$} & \makebox[3.15em][r]{$24\oounc{6}$}\,$\rightarrow$\,\makebox[3.15em][l]{$68\oounc{39}$} & \makebox[3.15em][r]{$\mathbf{25}\oounc{8}$}\,$\rightarrow$\,\makebox[3.15em][l]{$\mathbf{100}\oounc{1}$} \\
\texttt{scene} & \makebox[3.15em][r]{$0\oounc{0}$}\,$\rightarrow$\,\makebox[3.15em][l]{$0\oounc{0}$} & \makebox[3.15em][r]{$0\oounc{0}$}\,$\rightarrow$\,\makebox[3.15em][l]{$60\oounc{14}$} & \makebox[3.15em][r]{$\mathbf{82}\oounc{11}$}\,$\rightarrow$\,\makebox[3.15em][l]{$\mathbf{100}\oounc{1}$} & \makebox[3.15em][r]{$\mathbf{0}\oounc{0}$}\,$\rightarrow$\,\makebox[3.15em][l]{$\mathbf{100}\oounc{0}$} & \makebox[3.15em][r]{$15\oounc{4}$}\,$\rightarrow$\,\makebox[3.15em][l]{$10\oounc{3}$} & \makebox[3.15em][r]{$\mathbf{92}\oounc{23}$}\,$\rightarrow$\,\makebox[3.15em][l]{$\mathbf{100}\oounc{0}$} & \makebox[3.15em][r]{$\mathbf{51}\oounc{16}$}\,$\rightarrow$\,\makebox[3.15em][l]{$\mathbf{97}\oounc{4}$} & \makebox[3.15em][r]{$\mathbf{96}\oounc{9}$}\,$\rightarrow$\,\makebox[3.15em][l]{$\mathbf{100}\oounc{0}$} \\
\bottomrule
\end{tabular}%
}
\endgroup
\end{table}

\subsection{Offline-to-online RL}
\suppressfloats[t]

\noindent
\begin{minipage}[t]{0.57\linewidth}
\vspace{0pt}
\Cref{fig:o2o} and \Cref{tab:offline2online-eval} show offline-to-online RL performance of \oursplain{} on six OGBench tasks. \oursplain{} has the highest average success at the end of both phases and finishes within 95\% of the best method on all six tasks. Its average success rises from 67.1\% to 98.9\%, about 19 points above Value Flows and 26 above FQL in \Cref{tab:offline2online-eval}. BFN finishes second at 88.7\% but scores 45.4\% after offline training, below Value Flows. It edges ahead of \oursplain{} on \texttt{antmaze-large} but reaches only 90\% on the two cube tasks and 89\% on \texttt{humanoidmaze-medium}, and two of its eight seeds collapse on \texttt{puzzle-4x4}. RLPD, which learns only online, reaches 99--100\% on three tasks but stays near zero on the other three.
\end{minipage}\hfill
\begin{minipage}[t]{0.40\linewidth}
\vspace{0pt}
\centering
\includegraphics[width=\linewidth]{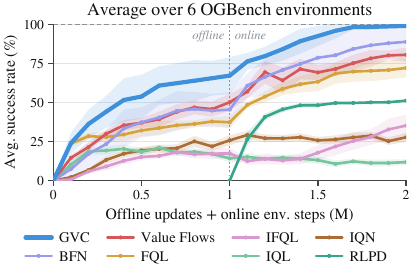}
\captionof{figure}{\footnotesize \textbf{Offline-to-online RL on OGBench.} Average success rate (\%) across six OGBench environments. The first $10^6$ steps are offline, followed by online finetuning for $10^6$ steps. We compare \oursplain{} with BFN, Value Flows, FQL, IFQL, IQL, IQN, and RLPD.}
\label{fig:o2o}
\end{minipage}

\subsection{Ablations}

\begin{figure}[H]
    \centering
    \includegraphics[width=\linewidth]{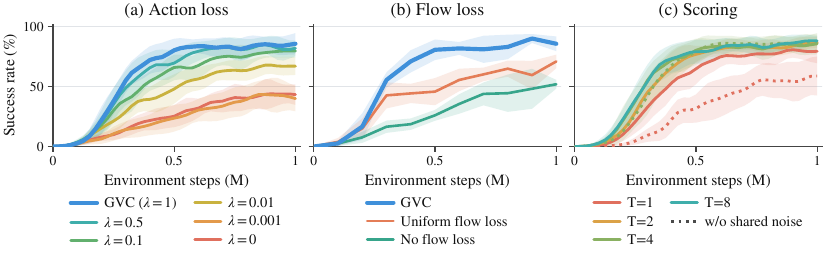}%
    \begin{subfigure}{0pt}\phantomsubcaption\label{fig:action-loss-ablation}\end{subfigure}%
    \begin{subfigure}{0pt}\phantomsubcaption\label{fig:tool-hang-cfm-loss-v}\end{subfigure}%
    \begin{subfigure}{0pt}\phantomsubcaption\label{fig:scoring-ablation}\end{subfigure}
    \caption{\footnotesize \textbf{Ablations on Tool Hang.} Success rate (\%) vs. environment steps. (a)~Action loss: varying flow loss coefficients, $\lambda_F$. (b)~Flow loss: \oursplain{} where the flow loss is weighted by the target return distribution $\widehat p_i$ (blue), uniform weighting across return bins (coral), and the flow loss removed (teal). (c)~Scoring: different $T$ values, with and without shared noise. Shaded regions show one standard deviation across seeds.} %
    \label{fig:ablations-toolhang}
\end{figure}

Our ablations look at the generative loss, which distinguishes \oursplain{} from a discriminative critic, as well as the impact of timestep sampling to approximate the action likelihood. First, \Cref{fig:action-loss-ablation} varies the coefficient $\lambda_F$ on $\mathcal L_{\mathrm{flow}}$ from its default of $1$ down to $0$. On Tool Hang, we find that the ordering is monotone in $\lambda_F$, and success drops from about 85\% at $\lambda_F=1$ to 80\% at $0.1$, 65\% at $0.01$, and 45\% either at $0.001$ or with the loss removed. Next, \Cref{fig:tool-hang-cfm-loss-v} looks at which part of the loss matters. On Tool Hang, we find that weighting all return bins uniformly---so that the flow loss trains the action model but no longer links actions to returns---recovers about 70\%, between the full loss at 85\% and no flow loss at about 50\%: roughly half the benefit comes from the generative supervision as such, and the rest from conditioning it on the return. Finally, \Cref{fig:scoring-ablation} varies the number of noise timesteps, $T$, used to score actions, as well as how the noise is sampled across action candidates. With noise shared across candidates, $T=1$ reaches about 80\% on Tool Hang, and using $T\geq2$ adds a few points; without sharing, $T=1$ falls to about 60\%. \Cref{app:ablations} shows results for the same set of ablations on the easier Square task, as well as ablations over the value-anchor weight $\lambda_V$, and a comparison of best-of-$N$ selection with a one-step actor distilled from critic gradients.

\section{Discussion} \label{sec:discussion}

We have presented \oursplain{}, a method for value learning that leverages a return-conditioned generative likelihood to implicitly model the return distribution. Our work builds off of recent work on generative classifiers and extends it to a domain where sample-efficiency is particularly important.

\paragraph{Limitations.}
\oursplain{} increases the theoretical compute costs of the critic, requiring $T$ forward passes to compute the likelihood, over $K$ bins, possibly over ensembled copies. In practice, we reduce this to only two forward passes per timestep and find that the compute costs are comparable to those of other performant distributional RL methods (\Cref{app:compute}), but this remains a consideration. Finally, in the form presented, \oursplain{} is factorized into two model components and requires discretizing the return space, both of which increase the implementation complexity.

\nonsubmissiononly{
\section{Acknowledgments}
We thank Armaan Abraham and Lars Ankile for helpful feedback on drafts of this manuscript.

This work was supported by an NSF CAREER award. This work was made possible by the Cloud TPUs provided through Google's TPU Research Cloud (TRC) and TPU Builder programs. This also used the Delta system at the National Center for Supercomputing Applications [award OAC 2005572] through allocation CIS260152 from the Advanced Cyberinfrastructure Coordination Ecosystem: Services \& Support (ACCESS) program, which is supported by U.S. National Science Foundation grants \#2138259, \#2138286, \#2138307, \#2137603, and \#2138296.
}

\nocite{dong2026tql}

\clearpage
\appendix
\section{Additional Ablations}
\label{app:ablations}

\Cref{sec:exp} reports our ablations on Tool Hang. This appendix adds the same ablations on Square (\Cref{app:ablations-square}), a sweep over the weight $\lambda_V$ of the value anchor (\Cref{app:value-anchor}), and a comparison of best-of-$N$ selection with a one-step actor trained by gradient ascent on $Q$ (\Cref{app:actor-distillation}). Unless stated otherwise, every run uses the settings of \Cref{tab:gvc-domain}.

\subsection{Ablations on Square}
\label{app:ablations-square}

\begin{figure}[H]
    \centering
    \includegraphics[width=\linewidth]{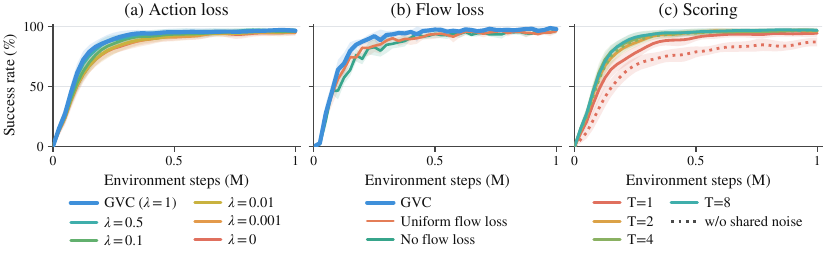}
    \caption{\footnotesize \textbf{Ablations on Square.} Success rate (\%) versus environment steps. (a)~Action loss: different flow-loss coefficients $\lambda_F$, compared with the default. (b)~Flow loss: \oursplain{} with the default flow loss weighted by the target return distribution $\widehat p_i$ (blue), uniform weighting across return bins (coral), and the flow loss removed (teal). (c)~Scoring: different $T$ values, with and without shared noise. Shaded regions show one standard deviation across seeds.}
    \label{fig:app-ablations}
\end{figure}

\Cref{fig:app-ablations} repeats the three ablations of \Cref{fig:ablations-toolhang} on Square, with the same settings and $8$ seeds per variant. Square is the easier task and separates the variants far less. Every flow-loss coefficient in (a) and every flow-loss variant in (b), including removing the loss, ends near $95\%$; the default learns slightly faster early on, but the curves otherwise nearly coincide. Scoring (c) tells the same story as on Tool Hang. With shared noise, $T=1$ learns a little more slowly than $T\geq2$ but catches up; without sharing, $T=1$ learns more slowly still and ends several points lower.

\subsection{Value-anchor coefficient}
\label{app:value-anchor}

\begin{figure}[htb]
\centering
\includegraphics[width=\linewidth]{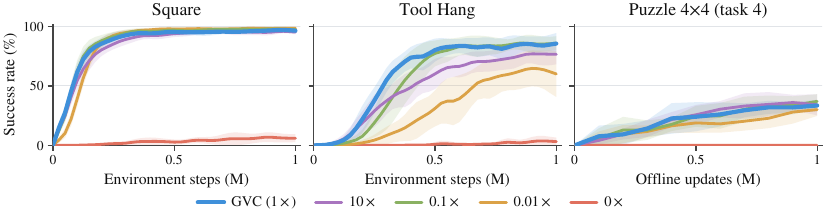}
\caption{\footnotesize \textbf{Value-anchor ablation on Square, Tool Hang, and \texttt{puzzle-4x4}.} Success rate versus environment steps, or offline updates on \texttt{puzzle-4x4} (\texttt{task4}), with the anchor coefficient $\lambda_V$ scaled to $10$, $0.1$, $0.01$, and $0$ times its default ($1\times$, \Cref{tab:gvc-domain}), $8$ seeds per setting.}
\label{fig:value-anchor-ablation}
\end{figure}

Value anchoring (Eq.~\eqref{eq:value-anchors}) fits $Q(s,a)$---the mean of the return distribution---to the TD target. \Cref{fig:value-anchor-ablation} varies its weight $\lambda_V$ on Square, Tool Hang, and offline \texttt{puzzle-4x4}, with everything else as in \Cref{tab:gvc-domain}. Without the anchor, success stays below $10\%$ on all three tasks. With it, Square is insensitive to the weight, and any non-zero value matches the default. Tool Hang is more sensitive but tolerates a wide range: $0.1\times$ ends near the default $1\times$ ($88\%$ vs.\ $86\%$), $10\times$ learns more slowly and ends at $77\%$, and $0.01\times$ is slower still ($57\%$). On \texttt{puzzle-4x4} the weight matters little once it is non-zero, and all four settings end between $30\%$ and $38\%$ after $10^6$ updates.

\subsection{Best-of-$N$ selection versus a distilled actor}
\label{app:actor-distillation}

\begin{figure}[H]
\centering
\includegraphics[width=\linewidth]{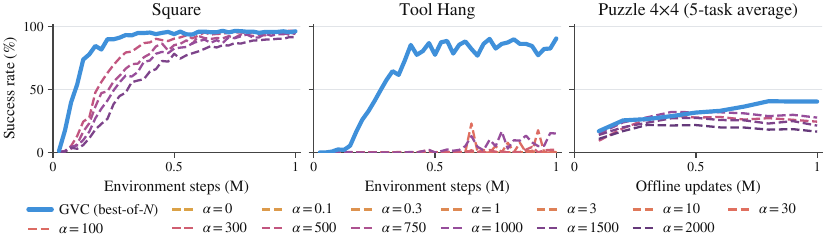}
\caption{\footnotesize \textbf{Best-of-$N$ selection versus a distilled one-step actor.} Success rate of \oursplain{} with its default best-of-$N$ extraction (solid) and with a one-step actor (shortcut model) trained by direct Q-gradients under \oursplain{}, for a range of distillation coefficients $\alpha$ (dashed). Left and center: online RL on Square and Tool Hang. Right: offline RL on \texttt{puzzle-4x4}, averaged over its five tasks. $\alpha=0$ removes the behavior anchor from the one-step actor.}
\label{fig:actor-distillation}
\end{figure}

\oursplain{} scores actions, so it can also drive an explicit actor that maximizes $Q$ by gradient ascent. Here we test the extraction recipe of FQL~\citep{park2025flowqlearning}: a one-step policy $\mu_\eta(s,\epsilon)$ that maps noise to an action is trained with
\begin{equation}
    \mathcal L_{\mathrm{distill}}(\eta)
    =\mathbb E_{s\sim\mathcal D,\,\epsilon\sim\mathcal N(0,I)}\!\left[
      \alpha\left\|\mu_\eta(s,\epsilon)-\mathrm{Flow}_\phi(s,\epsilon)\right\|_2^2
      -Q\!\left(s,\mu_\eta(s,\epsilon)\right)\right],
    \label{eq:distill-actor}
\end{equation}
where $\mathrm{Flow}_\phi(s,\epsilon)$ is the action obtained by integrating the BC flow $f_\phi$ from $\epsilon$ with $10$ Euler steps, $Q$ is the \oursplain{} critic, and its gradient with respect to the action is taken through the flow surrogate of Eq.~\eqref{eq:action-flow-error}. The BC flow actor is trained as before, and the one-step actor is used instead of best-of-$N$ selection. This follows FQL~\citep{park2025flowqlearning} where the $Q$ term is not normalized and other hyperparameters follow \Cref{tab:gvc-domain}. We sweep $\alpha$ over $\{0,0.1,0.3,1,3,10,30,100,300,1000\}$ on Tool Hang, $\{500,750,1000,1500\}$ on Square, and $\{500,1000,1500,2000\}$ on \texttt{puzzle-4x4}.

\Cref{fig:actor-distillation} shows the results. On Square, the distilled actor eventually matches the performance of best-of-$N$ by $10^6$ steps ($96\%$ for the best $\alpha$) but learns more slowly, reaching $78$--$90\%$ at $5\times10^5$ steps vs. $96\%$ for best-of-$N$. On Tool Hang, experiments do not find any coefficient that works: best-of-$N$ reaches $91\%$, while the distilled actors stay near zero. On \texttt{puzzle-4x4}, the distilled actor(s) peak at $\sim 30\%$ around $4\times10^5$ steps and then decline, while best-of-$N$ improves throughout to $40\%$.

Best-of-$N$ sampling only considers actions that have been sampled from the behavior flow, so it does not leave the support of the data. The distilled actor follows $\nabla_a Q$ instead, increasing the risk of instability. We find that only coefficients in the hundreds and above make progress, and on Tool Hang none does. The main results therefore use best-of-$N$ throughout. Extraction methods that use the action gradient while staying in support, such as adjoint matching~\citep{li2026q}, may narrow this gap in the future.

\section{Compute Cost}
\label{app:compute}

\oursplain{} spends more compute per scored action than a discriminative critic. Instead of predicting a value in a single forward pass, it evaluates the return-conditioned likelihood of the action under every return bin. In principle, the action model could be any generative model with an exact likelihood in a single forward pass, such as a discrete autoregressive or Gaussian model. Diffusion and flow-matching policies are the strongest architectures on the tasks we evaluate, however, so we use the same architecture for the critic and approximate the likelihood over noise timesteps as described in \Cref{ssec:flowlikelihood}.

We also reuse the policy architecture for the critic in every experiment, whereas most strong value-learning algorithms use a critic far smaller than the policy~\citep{dong2026tql}. We see this as a strength, since most prior methods cannot benefit from naively scaling the critic and in fact suffer from it~\citep{dong2026tql}. In practice, moreover, value learning is usually bottlenecked by environment interaction rather than by critic compute~\citep{dulacarnold2020empirical}.

We keep the cost down by predicting the action likelihood for all return bins in a single forward pass through a lightweight head, following~\citet{chen2024robust}. Most of the network capacity therefore goes into learning $p_\theta(a\mid s)$, and the head performs the final conversion to $p_\theta(a\mid s,z)$. Measured in forward evaluations (NFEs), \oursplain{} is then comparable to any diffusion or flow-based critic, such as Value Flows, that must integrate over timesteps.

\begin{figure}[htb]
\centering
\includegraphics[width=\linewidth]{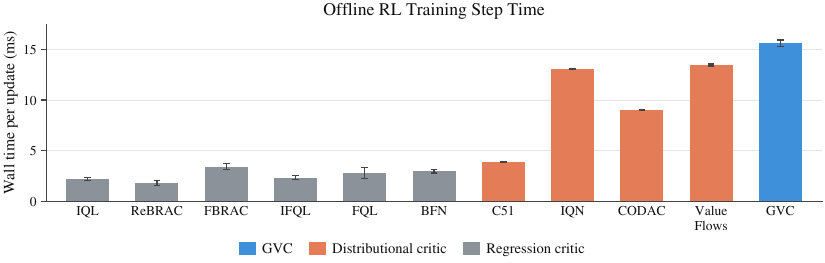}
\caption{\footnotesize \textbf{Offline RL training step time.} Wall-clock time per gradient update on \texttt{puzzle-4x4} on a single H100, mean and standard deviation over $3$ seeds.}
\label{fig:step-time}
\end{figure}

Concretely, \Cref{fig:step-time} reports the wall-clock time per gradient update for \oursplain{} and the offline baselines on \texttt{puzzle-4x4}, each run with its reference implementation and published hyperparameters on a single NVIDIA H100. \oursplain{} takes $15.6$\,ms per update, comparable to Value Flows ($13.5$\,ms) and IQN ($13.1$\,ms), the two distributional critics that also evaluate many return bins or quantiles per transition, and $5$--$9\times$ the cost of the scalar regression critics of IQL, ReBRAC, FBRAC, IFQL, and FQL ($1.8$--$3.4$\,ms).

\section{Implementation Details}
\label{app:implementation}

\subsection{\oursname{} hyperparameters}
\label{app:gvc-hparams}

\Cref{tab:gvc-common} lists the hyperparameters shared by all experiments and \Cref{tab:gvc-domain} lists those that differ by domain. The paragraphs below expand on both.

\begin{table}[H]
\centering
\small
\setlength{\tabcolsep}{5pt}
\caption{\footnotesize \textbf{Common \oursname{} hyperparameters.} Values shared by all experiments. Domain-specific values are given in \Cref{tab:gvc-domain}.}
\label{tab:gvc-common}
\begin{tabular}{@{}ll@{}}
\toprule
\textbf{Hyperparameter} & \textbf{Value} \\
\midrule
Optimizer & Adam \\
Learning rate & $3\times10^{-4}$ \\
Batch size & $256$ \\
Target network update rate & $0.005$ \\
Actor $f_\phi$ & MLP, $4$ hidden layers of width $512$, no layer normalization \\
Action model $g_\theta$ trunk & MLP, $3$ hidden layers of width $512$, layer normalization, GELU \\
Return-bin head & per-bin $512$-d embedding, one hidden layer of width $512$, linear output \\
Return prior $p_\psi$ & MLP on $s$, $3$ hidden layers of width $512$, layer normalization, GELU \\
Flow-time embedding (critic) & sinusoidal, $32$-d \\
Flow steps (Euler) & $10$ \\
Flow time for $\mathcal L_{\mathrm{flow}}$ and $\mathcal L_{\mathrm{actor}}$ & $\lambda\sim\mathcal U[0,1]$, one sample per transition \\
Scoring flow times $T$ & $4$, stratified on $[0.1,0.9]$, antithetic noise pair per time \\
Loss and prior weights & $\lambda_F=\lambda_R=1$, $\beta_{\mathrm{prior}}=0.1$ \\
Per-domain weights & $\beta_{\mathrm{ret}}$, $\lambda_V$ (\Cref{tab:gvc-domain}) \\
Update-to-data ratio (online) & $1$ \\
Offline gradient steps (OGBench) & $10^6$ \\
Online environment steps & $10^6$ \\
Evaluation episodes & $100$ \\
\bottomrule
\end{tabular}
\end{table}

\begin{table}[htb]
\centering
\small
\setlength{\tabcolsep}{4pt}
\caption{\footnotesize \textbf{\oursname{} hyperparameters.} $K$ is the number of return bins; $N$ is the number of candidates for best-of-$N$ selection; ``Critics'' gives the ensemble size and aggregation.}
\label{tab:gvc-domain}
\begin{tabular}{@{}lcccccc@{}}
\toprule
Domain & $\gamma$ & $K$ & $\beta_{\mathrm{ret}}$ & $\lambda_V$ & $N$ & Critics \\
\midrule
\multicolumn{7}{@{}l}{\textit{RoboMimic, online ($h=5$)}} \\
\texttt{lift}, \texttt{can}, \texttt{square} (MH) & $0.99$ & $5$ & $40$ & $0.15$ & $8$ & $2$, min \\
\texttt{tool-hang} (PH) & $0.99$ & $5$ & $40$ & $0.05$ & $8$ & $2$, min \\
\midrule
\multicolumn{7}{@{}l}{\textit{OGBench, offline ($h=1$)}} \\
\texttt{cube-double-play} & $0.995$ & $21$ & $40$ & $0.2$ & $8$ & $2$, mean \\
\texttt{cube-triple-play} & $0.995$ & $21$ & $10$ & $0.2$ & $8$ & $2$, mean \\
\texttt{puzzle-3x3-play-sparse} & $0.99$ & $51$ & $40$ & $0.2$ & $8$ & $2$, mean \\
\texttt{puzzle-4x4-play} & $0.99$ & $11$ & $40$ & $0.2$ & $8$ & $2$, mean \\
\texttt{puzzle-4x5-play} & $0.99$ & $21$ & $40$ & $0.5$ & $8$ & $2$, mean \\
\texttt{puzzle-4x6-play} & $0.99$ & $21$ & $40$ & $0.5$ & $8$ & $2$, mean \\
\texttt{scene-play-sparse} & $0.995$ & $11$ & $40$ & $0.05$ & $4$ & $2$, mean \\
\midrule
\multicolumn{7}{@{}l}{\textit{OGBench, offline-to-online ($h=1$)}} \\
\texttt{antmaze-large-navigate} & $0.99$ & $21$ & $40$ & $0.2$ & $4$ & $2$, mean \\
\texttt{humanoidmaze-medium-navigate} & $0.995$ & $21$ & $40$ & $0.2$ & $8$ & $2$, mean \\
\texttt{cube-double-play} & $0.995$ & $21$ & $40$ & $0.2$ & $4$ & $2$, mean \\
\texttt{cube-triple-play} & $0.995$ & $21$ & $40$ & $0.2$ & $4$ & $2$, mean \\
\texttt{puzzle-4x4-play} & $0.99$ & $51$ & $40$ & $0.2$ & $8$ & $2$, mean \\
\texttt{scene-play} & $0.995$ & $51$ & $40$ & $0.05$ & $8$ & $2$, mean \\
\bottomrule
\end{tabular}
\end{table}

\paragraph{Networks.}
The actor $f_\phi$ is the flow policy of FQL and QC, an MLP on $(s,x_\lambda,\lambda)$. The action model $g_\theta$ takes the same inputs, with $\lambda$ passed through a sinusoidal embedding. The action model conditions on the return bin only after the trunk, and we add a learned embedding of each bin to the trunk features, and apply a small head, shared across bins. One forward pass therefore covers every bin, and these embeddings are the only bin-specific parameters. The return prior $p_\psi(z\mid s)$ is a separate MLP on $s$ with the same trunk architecture and $K$ output logits. For ensembling (as denoted in \Cref{tab:gvc-domain}), we train independent copies of the action model and prior. With ``mean'' we average the two return distributions bin by bin. With ``min'' we keep whichever one has the lower mean. This combined distribution is what ranks actions and what $\bar p$ denotes in \Cref{alg:gvc}. Target networks are exponential moving averages of the online ones.

\paragraph{Scoring.}
We estimate Eq.~\eqref{eq:action-flow-error} at $T=4$ flow times, stratified over $[\lambda_{\min},\lambda_{\max}]$ (typically $\lambda_{\min}=0.1$, $\lambda_{\max}=0.9$), with an antithetic noise pair $(\epsilon,-\epsilon)$ at each time. Scoring a candidate therefore costs $2T$ forward passes of $g_\theta$, each of which returns all $K$ bins. The same noise and times are reused across bins and across the $N$ candidates at a state. We use this estimator everywhere the critic is used for scoring. The training loss $\mathcal L_{\mathrm{flow}}$ instead draws one fresh $(\epsilon,\lambda)$ per transition, with $\lambda\sim\mathcal U[0,1]$. \Cref{fig:scoring-ablation} varies $T$ while keeping the antithetic pairs; its variants with non-shared noise draw independent noise and times for each candidate. For RoboMimic \texttt{lift}, \texttt{can}, and \texttt{square}, as well as OGBench \texttt{cube} environments, we find that scoring over each action dimension separately produces more accurate likelihood estimates.

\paragraph{Return bins and loss weights.}
We discretize returns into $K$ uniformly spaced bins over $[z_1,z_K]$. This is typically set from the analytical minimum and maximum return given the environment reward and discount. The projection $\Pi_{\mathcal Z}$ in \Cref{alg:gvc} is defined in~\citet{bellemare2017distributional} and the shifted atoms $r+\gamma m z_j$ are first clipped to $[z_1,z_K]$. The mass is then split between the two closest bins by linear interpolation. To keep the predicted distribution from collapsing to a single bin, we smooth the projected target $\widehat p$ with a Gaussian kernel similar to HL-Gauss~\citep{imani2018improving,farebrother2024stop}.

For Eq.~\eqref{eq:return-posterior}, we weight these two terms separately, with $\beta_{\mathrm{ret}}$ on $\log p_\theta(a\mid s,z_i)$ and $\beta_{\mathrm{prior}}$ on $\log p_\psi(z_i\mid s)$ (\Cref{tab:gvc-common}). Unless otherwise noted, we set $\lambda_F=\lambda_R=1$ and pick $\lambda_V$ for each domain (\Cref{tab:gvc-domain}). \Cref{fig:action-loss-ablation} varies $\lambda_F$, and \Cref{fig:value-anchor-ablation} varies $\lambda_V$.

\paragraph{Actor and extraction.}
In all settings, the actor is the behavior-cloning flow of Eq.~\eqref{eq:actor-loss}, trained on the actions in the batch (dataset actions offline and replay actions online). We extract a policy by best-of-$N$ sampling: sample $N$ actions from the flow with $10$ Euler steps, score them with the critic, and execute the one with the highest $Q(s,a)$.

We re-sample at every environment step, or after each chunk for RoboMimic. We choose $a'$ for the TD target the same way and use the online critic to rank the candidates and the target critic to score the selected one. We use direct Q-gradients only in the ablation in \Cref{app:actor-distillation}.

\subsection{Training and evaluation protocol}
\label{app:protocol}

\paragraph{RoboMimic.}
We follow the QC recipe. There is no offline phase: updates start after $5{,}000$ environment steps with an update-to-data (UTD) ratio of $1$, and each batch is a 50/50 mix of demonstration transitions (multi-human for Lift, Can, and Square; proficient-human for Tool Hang) and replay transitions. At inference, the policy predicts chunks of $h=5$ actions and executes them open-loop. As in QC, the critic scores whole chunks. A transition is $(s_t,a_{t:t+h},\sum_{k<h}\gamma^k r_{t+k},s_{t+h})$ and the target uses $\gamma^h$ in place of $\gamma$. We use a sparse reward of $-1$ at every step until success and $0$ at success. Episodes end at success, so a chunk that ends the episode has $m=0$, and time-outs still bootstrap. We use $3$ seeds and evaluate for $100$ episodes every $25{,}000$ steps. The ablations in \Cref{sec:exp} and \Cref{app:ablations-square} use the Square and Tool Hang settings of \Cref{tab:gvc-domain} with $8$ seeds.

\paragraph{OGBench.}
We follow the Value Flows recipe. Offline results use $10^6$ gradient steps on the offline dataset with $h=1$, whereas offline-to-online results follow the same recipe but add an additional $10^6$ environment steps at an update-to-data (UTD) ratio of $1$, with updates resuming after $1{,}000$ online steps. The sparse \texttt{scene} and \texttt{puzzle-3x3} variants follow QC. We relabel the dataset rewards to $-1$, except at steps where the task is complete, which get $0$. For offline-to-online we use \texttt{task1} of \texttt{antmaze-large-navigate}, \texttt{humanoidmaze-medium-navigate}, and \texttt{cube-triple-play}, \texttt{task2} of \texttt{cube-double-play} and \texttt{scene-play}, and \texttt{task4} of \texttt{puzzle-4x4-play}, all with the default rewards. The replay buffer starts out as the offline dataset. Online transitions are added to it as they are collected, and batches are always sampled uniformly from the entire buffer. We use $8$ seeds and evaluate for $100$ episodes every $10^5$ steps; \Cref{tab:offline-eval-full} reports the final checkpoint.

\paragraph{Tables.}
Entries in \Cref{tab:offline-eval-full,tab:offline2online-eval} are means $\pm$ sample standard deviations over $8$ seeds, in percentage points.

\paragraph{Learning curves.}
In \Cref{fig:robomimic-online} we smooth each seed's curve with a Gaussian kernel (bandwidth $2\%$ of the observations) and then take the mean and sample standard deviation across seeds. In \Cref{fig:o2o} the \oursplain{} and BFN curves are unsmoothed averages of the six per-task means and their bands are averages of the six per-task standard deviations, with $8$ seeds per task. The other baseline curves and bands are taken from Value Flows~\citep{dong2026valueflows}, except RLPD, which we ran ourselves. \Cref{fig:ablations-toolhang,fig:app-ablations,fig:value-anchor-ablation} use $8$ seeds per variant, and shading shows one standard deviation.

\subsection{Baseline tuning}
\label{app:baseline-tuning}

\paragraph{Reference results.}
For offline OGBench results on \texttt{puzzle-\allowbreak{}4x4-\allowbreak{}play}, \texttt{cube-\allowbreak{}double-\allowbreak{}play}, and \texttt{cube-\allowbreak{}triple-\allowbreak{}play}, we report the baseline numbers from Value Flows~\citep{dong2026valueflows}. We run the remaining baselines ourselves and tune them as described below. For offline-to-online OGBench results, we report the numbers from Value Flows~\citep{dong2026valueflows} for all baselines except BFN.

\paragraph{Tuning procedure.}
We follow the tuning protocol of Value Flows~\citep{dong2026valueflows}: each hyperparameter is swept on the domain's default task over the values in \Cref{tab:baseline-tuning}. The BC policy is Gaussian with four hidden layers of width $512$, and CODAC's conservative penalty is fixed at $0.1$.

\paragraph{BFN.}
BFN shares the actor, discount, critic ensemble, and training schedule of \oursplain{} (\Cref{tab:gvc-domain}) and replaces the generative critic with the regression critic of QC~\citep{li2025actionchunking}.

\begin{table}[H]
\centering
\small
\setlength{\tabcolsep}{4pt}
\caption{Hyperparameter search spaces for tuning our baselines, following Value Flows~\citep{dong2026valueflows}. BC denotes behavioral cloning; BCFM denotes bootstrapped conditional flow matching.}
\label{tab:baseline-tuning}
\begin{tabular}{@{}lll@{}}
\toprule
Method & Hyperparameter & Search values \\
\midrule
IQL & AWR inverse temperature & $\{0.3,1,3,10\}$ \\
ReBRAC & Actor BC coefficient & $\{0.003,0.01,0.03,0.1,0.3,1\}$ \\
 & Critic BC coefficient & $\{0,0.001,0.01,0.1\}$ \\
IFQL & Action candidates & $\{32,64,128\}$ \\
FBRAC & BC coefficient & $\{1,3,10,30,100,300\}$ \\
FQL & BC distillation coefficient & $\{3,10,30,100,300,1000\}$ \\
C51 & Distributional atoms & $\{51,101\}$ \\
IQN, CODAC & Quantile-loss $\kappa$ & $\{0.7,0.8,0.9,0.95\}$ \\
CODAC & BC coefficient & $\{100,300,1000,3000,10000,30000\}$ \\
Value Flows & BCFM coefficient $\lambda$ & $\{0.3,0.5,1,3,10\}$ \\
 & Confidence temperature $\tau$ & $\{0.01,0.03,0.1,0.3,1,3\}$ \\
 & Online distillation coefficient & $\{10,30,100,300\}$ \\
\bottomrule
\end{tabular}
\end{table}

\end{document}